\documentclass{article}
\usepackage{mysty,spconf,amsmath,graphicx}

\usepackage{enumitem}
\usepackage{cleveref}
\setlist{nosep, leftmargin=14pt}

\usepackage{mwe} 

\title{Neural Blind Deconvolution for Image Scanning Microscopy}
\name
 {Eliot Guez$^1$, Nathan Buskulic$^1$, Lisa Cuneo$^{2,3}$, Luca Calatroni$^{1,3}$, Giuseppe Vicidomini$^3$}
\address{%
$^{1}$ MaLGa Center, DIBRIS, University of Genoa, Italy,\\
$^{2}$ CSLM, Italian Institute of Technology, Italy,
$^{3}$ MMS, Italian Institute of Technology, Italy
}

\begin{document}
%
\maketitle
\begin{abstract}

Image scanning microscopy (ISM) acquires one scanned image per detector element, which are then fused into a single image by multi-image deconvolution to obtain a super-resolved image. Its accuracy, however, depends on point spread function (PSF) models that sample- and system-induced aberrations often invalidate. Since differences between the scanned images encode aberration information, we propose a training-free blind Deep Image Prior (DIP) framework that jointly estimates the specimen and the detector-dependent PSFs. The PSFs are parametrised through the detector geometry and shared excitation and emission PSFs, reducing the unknowns while preserving the physical ISM model, and a Gaussian pre-initialisation stabilises the blind optimisation. On experimental tubulin data, blind DIP recovers finer filamentous structures than multi-image deconvolution and DIP with fixed theoretical PSFs, while adapting the PSFs to the measurements.
\end{abstract}
\section{Introduction}
\label{sec:intro}

Image scanning microscopy (ISM) replaces the single-element detector of confocal laser-scanning microscopy with a detector array to produce super-resolved images with a high signal-to-noise ratio (SNR) \cite{colin1988super}.
The array elements simultaneously record scanned images, each containing complementary spatial information owing to the different positions of the elements in the detection plane. Each element acts as a small pinhole, providing the resolution benefits of narrow-pinhole confocal detection, while neighboring elements collect otherwise rejected photons, preserving a high photon budget. Multi-image deconvolution (MID) combines these images using their detector-dependent PSFs ~\cite{zunino2023reconstructing}; extending its forward model to account for a three-dimensional specimen also enables optical sectioning \cite{Zunino2025}. These PSFs are often derived from theoretical models that may not capture experimental aberrations. Crucially, differences among the scanned images encode information about these aberrations \cite{Fersini2025}, providing additional constraints for PSF estimation and motivating blind deconvolution.

We consider here a two-dimensional blind deconvolution setting where the unknown sample $\xv\in\mathbb{R}_+^n$ gives rise to $N_{ch}=25$ observations $\{\yv_i\}_{i=1}^{N_{ch}}$, $\yv_i\in\mathbb{N}_0^n$, through related but distinct convolution kernels (i.e., detector-dependent PSFs) $\{\hv_i\}_{i=1}^{N_{ch}}$, $\hv_i\in\mathbb{R}_+^k$, where $n$ and $k$ denote the number of pixels in the image and kernel domains, respectively.
Since the SPAD array detects individual photons without readout noise, we model the observations with Poisson statistics, i.e.\ $\yv_i\sim\operatorname{Poisson}(\hv_i * \xv)$. As unknown aberrations can cause the experimental PSFs to differ from their theoretical models, both $\xv$ and $\{\hv_i\}_{i=1}^{N_{ch}}$ have to be recovered from $\{\yv_i\}_{i=1}^{N_{ch}}$, leading to a \textit{blind} inverse problem. Its set of solutions can, however, be restricted by exploiting the (possibly estimated) detector geometry, which determines how the kernels are related to one another.
%
%
Furthermore, a real dataset pairing ISM measurements with ground-truth images and PSFs is not easy to obtain for training. We therefore turn to unsupervised reconstruction and, more precisely, to SelfDeblur~\cite{ren2020neural}, a blind extension of the Deep Image Prior (DIP)~\cite{ulyanov2018deep}. In the non-blind setting, DIP has proved flexible and effective in practice~\cite{baguer2020computed,darestani2021accelerated}, and in some regimes it comes with theoretical convergence and recovery guarantees~\cite{buskulic2024convergence,buskulic2024recovery}.
Blind problems, however, raise additional challenges~\cite{campisi2017blind}. In deconvolution, even without noise, the factorisation $\yv=\xv\ast\hv$ is not unique: any rescaling $(c^{-1}\xv,c\hv)$ with $c>0$ or opposite translations $(T(\xv,\nu),T(\hv,-\nu))$, where $T(\cdot,\nu)$ shifts its input by $\nu$, yields the same observation. A classical example in single-image blind deconvolution is the trivial no-blur solution $(\yv,\delta)$, with $\delta$ a Dirac mass. A key issue when adapting DIP to the blind setting is therefore to select robust solutions among these equivalent ones.


In this work, we develop a DIP-based approach for blind ISM reconstruction that jointly estimates the underlying image and the set of convolution kernels. The optical ISM model expresses each kernel $\hv_i$ as the product of a shared excitation PSF and a detection PSF derived from the shared emission PSF, shifted according to the position of detector element $i$. We incorporate this physical relationship into the optimisation, reconstructing all kernels $\{\hv_i\}_{i=1}^{N_{ch}}$ from the two shared excitation and emission PSFs. This reduces the number of unknowns and restricts the set of admissible solutions. We further introduce a pre-initialisation strategy for the kernel-generating networks, which stabilises the blind optimisation and makes it robust to translation ambiguities. Experiments on real ISM data show promising qualitative results.

\section{Image Scanning Microscopy}



To describe precisely the underlying ISM image acquisition model, we set all detector elements to share the same geometry, described by a common binary spatial detection response $p$, assuming uniform detection efficiency across the active area. Each element’s response is therefore a shifted copy, $p_i(\mathbf r)=p(\mathbf r-\mathbf r_i)$, where $\mathbf r_i$ is the shift vector describing the position of element $i$ in the image plane. The corresponding detection and effective PSFs read
\begin{equation}
\hv_{\mathrm{det},i}=\hv_{\mathrm{em}}*\pv_i,\qquad
\hv_i=\hv_{\mathrm{exc}}\cdot \hv_{\mathrm{det},i},
\label{eq: ISM PSFs}
\end{equation}
where $\hv_{\mathrm{exc}}$ and $\hv_{\mathrm{em}}$ are the excitation and emission PSFs. Each effective PSF $\hv_i$ is narrower than $\hv_{\mathrm{em}}$, the PSF of a diffraction-limited widefield system, providing the basis for ISM’s super-resolution capability \cite{colin1988super}. 

Importantly, once the detector geometry is fixed, all $N_{ch}$ effective PSFs are determined by only two functions, $\hv_{\mathrm{exc}}$ and $\hv_{\mathrm{em}}$. In practice, all quantities are sampled on the kernel grid. Denoting by $\mathbf \hv_{\mathrm{exc}},\mathbf \hv_{\mathrm{em}},\mathbf p_i\in\mathbb R_+^k$ the discretised PSFs and apertures, \eqref{eq: ISM PSFs} becomes $\hv_i=\mathbf h_{\mathrm{exc}}\odot(\mathbf h_{\mathrm{em}}*\mathbf p_i)$, with $\odot$ the entrywise product. Since $\mathbf p_i=T(\mathbf p,\mathbf r_i)$ is a shifted copy of the discretised aperture $\mathbf p$ of a single element, the set $\{\mathbf p_i\}_{i=1}^{N_{ch}}$ is fully determined by the element shape and the positions $\mathbf r_i$, expressed in pixel units.

When $\{\hv_i\}_{i=1}^{N_{ch}}$ are known, an estimate of the desired object $\xv$ can be found by solving
\begin{equation}
\min_{\xv\geq 0}\;\sum_{i=1}^{N_{ch}} \mathrm{KL}\big(\yv_i,\hv_i*\xv\big),
\quad
\mathrm{KL}(\yv,\mathbf z)=\Big\langle \yv\log\frac{\yv}{\mathbf z}-\yv+\mathbf z,\mathbf 1\Big\rangle,
\label{eq: multi-image deconvolution}
\end{equation}
where the KL$(\cdot,\cdot)$ denotes the Kullback-Leibler divergence. Problem~\eqref{eq: multi-image deconvolution} is typically solved via Richardson--Lucy scheme extended to multiple images ~\cite{zunino2023reconstructing}, with $\mathbf h_{\mathrm{exc}}$ and $\mathbf h_{\mathrm{em}}$ calculated using vectorial diffraction theory. In the blind setting, these two PSFs become unknowns, and the discretised model above provides the physical structure on which our method builds in the next section.

\section{Deep Parametrisation for blind ISM reconstruction}

Deep Image Prior (DIP)~\cite{ulyanov2018deep} solves inverse problems without training data by relying on the implicit regularisation of a neural network rather than on an explicit regulariser. The unknown is parametrised as the output $\xvhat=\gv_\xv(\zv_\xv,\thetav_\xv)$ of a network $\gv_\xv:\R^d\times\R^p\to\R^n$, with fixed random input $\zv_\xv\in\R^d$ and weights $\thetav_\xv\in\R^p$. For a known kernel $\hv$ and a data-fidelity term $\mathcal L:\R^n\times\R^n\to\R_+$, the weights are computed by solving
\begin{equation}
\min_{\thetav_\xv}\ \mathcal L\big(\yv,\hv*\gv_\xv(\zv_\xv,\thetav_\xv)\big).
\label{eq:DIP}
\end{equation}


Several extensions of DIP to blind problems exist~\cite{ren2020neural,bredell2023wiener,zhang2024blind}. The closest to our purposes is SelfDeblur~\cite{ren2020neural}, which parametrises the kernel by a second network, $\hvhat=\gv_\hv(\zv_\hv,\thetav_\hv)$, and jointly solves
\begin{equation}
\min_{\thetav_\xv,\thetav_\hv}\ \mathcal L\big(\yv,\gv_\hv(\zv_\hv,\thetav_\hv)*\gv_\xv(\zv_\xv,\thetav_\xv)\big).
\label{eq:SelfDeblur}
\end{equation}

In ISM, the $N_{ch}$ deconvolution problems are coupled through three unknowns: the object $\xv$ and the PSFs $\mathbf h_{\mathrm{exc}}$, $\mathbf h_{\mathrm{em}}$. We therefore adapt SelfDeblur by parametrising each unknown with its own network, $\xvhat=\gv_\xv(\thetav_\xv)$, $\hvhat_{\mathrm{exc}}=\gv_{\mathrm{exc}}(\thetav_{\mathrm{exc}})$ and $\hvhat_{\mathrm{em}}=\gv_{\mathrm{em}}(\thetav_{\mathrm{em}})$, omitting the fixed inputs for simplicity. The object network has a non-negative output layer, while the two kernel networks share the same architecture and map onto the simplex $\Delta_k=\{\hv\in\R_+^k:\langle\hv,\mathbf 1\rangle=1\}$ through a final softplus followed by $\ell_1$ normalisation. With the spatial detection response $\pv_i$ fixed by the detector geometry (Section~2), we solve
\small
\begin{equation}\label{eq:DIP-ISM}\tag{DIP-ISM}
\begin{aligned}
&\min_{\thetav_\xv,\thetav_{\mathrm{exc}},\thetav_{\mathrm{em}}}\ \sum_{i=1}^{N_{ch}}\mathrm{KL}\big(\yv_i,\hvhat_i*\gv_\xv(\thetav_\xv)+\varepsilon\mathbf 1\big)+R(\hvhat_{\mathrm{exc}})+R(\hvhat_{\mathrm{em}}),\\
&\ \text{s.t.}\quad \hvhat_i=\hvhat_{\mathrm{exc}}\odot\big(\hvhat_{\mathrm{em}}*\mathbf p_i\big),
\end{aligned}
\end{equation}
\normalsize
where KL is defined in~\eqref{eq: multi-image deconvolution}, $R(\hv)=\frac{\lambda}{2}\| \nabla \hv\|_2^2$ is a Tikhonov regulariser stabilising the kernel estimation, and $\varepsilon>0$ is a small offset. Non-negativity and normalisation hold by construction, so~\eqref{eq:DIP-ISM} is unconstrained in the weights. Note that such normalisation removes the scale ambiguity problem, while preserving the relative intensities of the channels, which carry information on the detector geometry.

The objective of~\eqref{eq:DIP-ISM} can be differentiated with respect to the weights by backpropagation, and we minimise it with Adam. We stress that this only guarantees well-defined, stable gradients: since~\eqref{eq:DIP-ISM} is non-convex, no convergence guarantee is claimed, and the analyses of~\cite{buskulic2024convergence,buskulic2024recovery} do not directly cover the blind setting.


Note that the proposed approach couples the $N_{ch}$ kernels through the detector geometry, thus retaining physical knowledge on the excitation and emission PSFs that would be lost by estimating each $\hvhat_i$ independently, e.g.\ by means of $N_{ch}$ separate kernel networks.



\subsection{On the choice of the initialisation}




As mentioned in Section~\ref{sec:intro}, blind problems are faced with sets of ambiguous solutions. Notably, SelfDeblur, and vanilla~\eqref{eq:DIP-ISM} are very sensitive to finding translated solutions, which depend on the initialization. To mitigate this, before starting the optimization of~\eqref{eq:DIP-ISM}, we train for a few steps $\gv_\hv$ to output a centered isotropic Gaussian kernel of chosen variance $\sigma$. The goal is simply to give the optimization process a physically plausible starting point and, we hope, stay close to. This simple procedure proves to be very effective to guarantee centered solutions in practice, for various noise levels, initialisations and choices of $\sigma$.

Contrary to other existing stabilization methods in the litterature~\cite{bredell2023wiener,zhang2024blind}, this method does not necessitate any training set and works directly at the output level of the network making it very simple to use. Furthermore, pretraining the weights to meet an output also changes the Jacobian of $\gv_\hv$ to make a few of its singular values much more important than others, which does help at the beginning of the optimization to keep the solution close to its pretrained state.

\begin{figure*}[t!]
    \centering
    \includegraphics[width=0.88\linewidth]{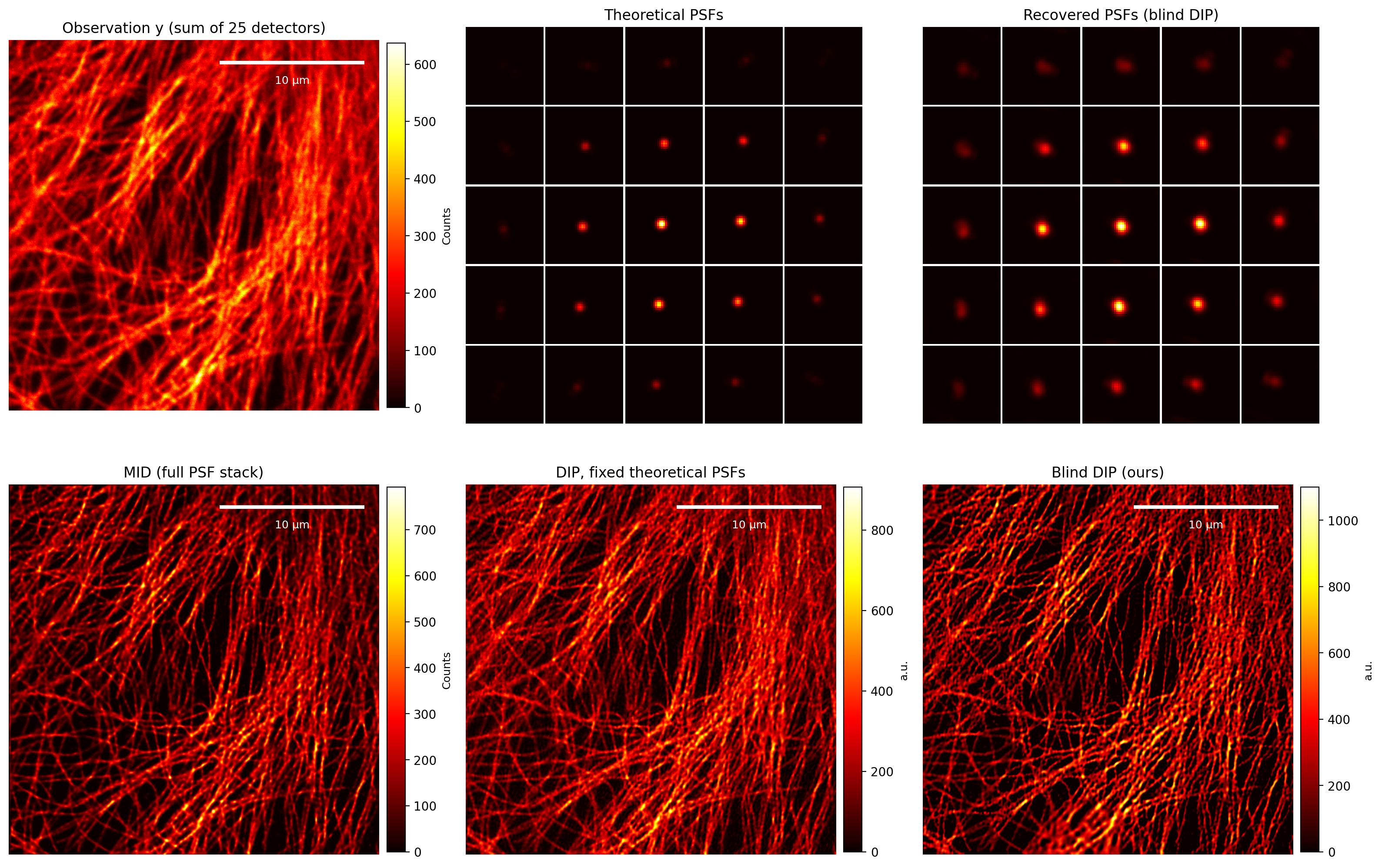}
    \caption{Comparison on experimental tubulin ISM data. Top row, from left to right: sum of the 25 detector observations, theoretical detector-dependent PSFs, and PSFs estimated by blind DIP. Bottom row: multi-image deconvolution (MID) with the full theoretical PSF stack, DIP with fixed theoretical PSFs, and the proposed blind DIP reconstruction. Image intensities are displayed with separate color scales.}
    \label{fig:method_comparison}
\end{figure*}

\section{Numerical Results}



We evaluated the proposed method from data obtained from a custom ISM system equipped with a $5\times5$ SPAD detector array \cite{Le2026}. 
We chose for~\eqref{eq:DIP-ISM} to reconstruct the specimen $\xv$ using $\gv_\xv$ as a convolutional skip network, following the architecture used in SelfDeblur~\cite{ren2020neural}, while $\gv_\hv$ was taken as a two-layer fully connected network with ReLU6 activation, followed by Softplus and $L_1$ normalization. We optimize jointly the three networks by doing $10^3$ Adam steps, with image and kernel learning rates of $10^{-2}$ and $10^{-4}$ respectively, to minimize $\mathcal{L}$.
For comparison, we used MID~\cite{zunino2023reconstructing}, which reconstructs the image through a multi-image Richardson--Lucy scheme using the 25 detector-dependent PSFs $\hv_i$ constructed from excitation and emission PSFs calculated using vectorial diffraction theory. We also evaluated a non-blind DIP approach using the same fixed theoretical PSFs, providing a baseline for assessing the benefit of joint image and PSF estimation.


\subsection{Qualitative comparison}
%

\Cref{fig:method_comparison} compares the conventional confocal image (obtained by summing all the scanned images), the theoretical and estimated effective PSFs, and the three image reconstructions.  All three reconstruction methods resolve the main network of tubulin filaments. Compared with MID and DIP with fixed theoretical PSFs, blind DIP produces visually more pronounced fine filament structures, which may reflect its ability to account for optical aberrations affecting the acquisition. 

\begin{figure}
    \centering
    \includegraphics[width=\linewidth]{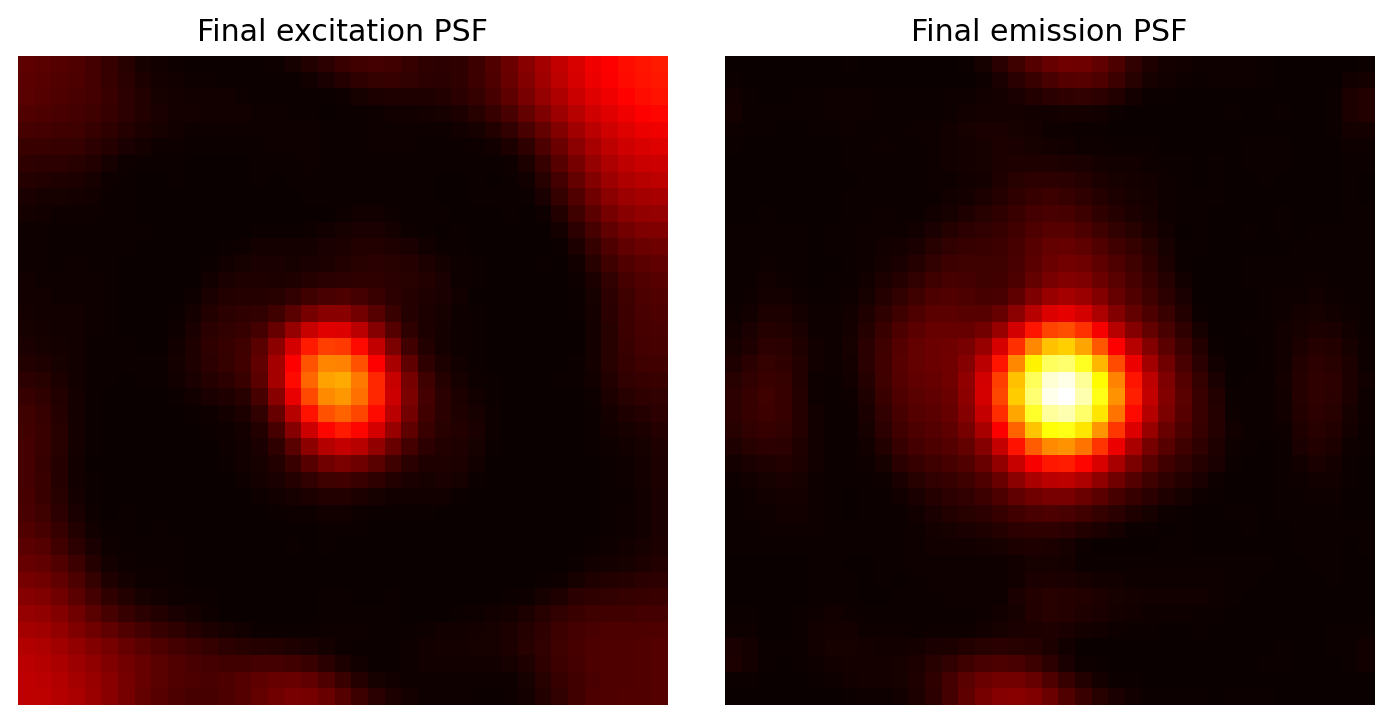}
    \caption{Latent excitation (left) and emission (right) PSFs estimated jointly with the specimen by the proposed blind DIP approach on experimental tubulin data. Both exhibit a localized central peak, with additional diffuse and peripheral intensity.}
    \label{fig:final_psf}
\end{figure}

The recovered effective PSFs differ from the theoretical ones in their apparent extent and spatial distribution (\cref{fig:method_comparison}). A main advantage of our~\eqref{eq:DIP-ISM} method is that we recover the underlying excitation and emission PSFs which allow a much finer analysis of the optical system. For example, in \cref{fig:final_psf} the estimated excitation PSF has a diagonally elongated central lobe, consistent with astigmatism, a typical optical aberration that could then be corrected. In these estimated PSFs, the peripheral intensity on the border is not optically plausible and probably reflects an artifact of the blind estimation. However, these peripheral artefacts are largely suppressed in $\hv_i$ by the pointwise multiplication of the spatially localised excitation and detection PSFs. 


\section{Conclusion}

We presented a blind DIP approach for jointly estimating the specimen image and the excitation and emission PSFs for ISM imaging. By constructing each detector-dependent effective PSF from the shared excitation and emission PSFs and the detector geometry, the method incorporates the physical relationships among the kernels.  On experimental tubulin data, the resulting reconstruction reveals fine filament details and estimates PSFs that differ from the theoretical model. These qualitative results support further investigation of blind DIP for ISM, including the stability and physical interpretation of the recovered PSFs and quantitative validation on suitable reference measurements.

\footnotesize
\section{Acknowledgments}
\label{sec:acknowledgments}

The work of EG, NB, Lu C was supported by the funding received from the European Research Council (ERC) Starting project MALIN under the European Union’s Horizon Europe programme (grant 101117133). The work of Li C was partially supported by the Teaching Science to Computer Flagship program of the Italian Institute of Technology. This work represents only the view of the authors. The European Commission and the other organisations are not responsible for any use that may be made of the information it contains. 

\normalsize

\bibliographystyle{IEEEbib}
\bibliography{references}

\end{document}